\documentclass[a4paper]{svproc}
\usepackage{multirow}
\usepackage{hyperref} 
\usepackage{booktabs} % For prettier tables
\usepackage{makecell}
\usepackage{caption}
\usepackage{tabularx}
\usepackage{amsmath,graphicx}
\usepackage{xcolor}
\usepackage{tabularx}
 \usepackage{pdflscape}
\usepackage{url}

\begin{document}
\mainmatter              % start of a contribution
\title{LRDDv2: Enhanced Long-Range Drone Detection Dataset with Range Information and Comprehensive Real-World Challenges}
\titlerunning{LRDDv2}  % abbreviated title (for running head)
\author{Amirreza Rouhi, Sneh Patel, Noah McCarthy, Siddiqa Khan, Hadi Khorsand, Kaleb Lefkowitz, David K.Han}
\authorrunning{Rouhi et al.} % abbreviated author list (for running head)

\institute{Department of Electrical and Computer Engineering, Drexel University, Philadelphia, PA, USA,\\
{\tt\small {\{ar3755, sp3568, nm999, sik33, hk852,kjl352, dkh42\}}@drexel.edu} 
}

\maketitle              % typeset the title of the contribution

\begin{abstract}

The exponential growth in Unmanned Aerial Vehicles (UAVs) usage underscores the critical need of detecting them at extended distances to ensure safe operations, especially in densely populated areas. Despite the tremendous advances made in computer vision through deep learning, the detection of these small airborne objects remains a formidable challenge. While several datasets have been developed specifically for drone detection, the need for a more extensive and diverse collection of drone image data persists, particularly for long-range detection under varying environmental conditions. We introduce here the Long Range Drone Detection (LRDD) Version 2 dataset, comprising 39,516 meticulously annotated images, as a second release of the LRDD dataset released previously. The LRDDv2 dataset enhances the LRDDv1 by incorporating a greater variety of images, providing a more diverse and comprehensive resource for drone detection research. What sets LRDDv2 apart is its inclusion of target range information for over 8,000 images, making it possible to develop algorithms for drone range estimation. Tailored for long-range aerial object detection, the majority of LRDDv2's dataset consists of images capturing drones with 50 or fewer pixels in 1080p resolution. For access to the complete Long-Range Drone Detection Dataset (LRDD)v2, please visit \\
\href{https://research.coe.drexel.edu/ece/imaple/lrddv2/}{https://research.coe.drexel.edu/ece/imaple/lrddv2/}.

\end{abstract}
\section{Introduction}

The rapid integration of Unmanned Aerial Vehicles (UAVs) into commercial applications has highlighted the critical need for advanced detection systems capable of accurately sensing these small objects over considerable distances. These systems are essential for ensuring the safe navigation of UAVs, especially in environments where the risk of collision with other manned or unmanned aircraft is significant. Many of these detection systems rely on deep learning algorithms, making the quantity and diversity of the datasets used for training crucial to their effectiveness \cite{rouhi2024overview}. However, existing datasets often lack comprehensive coverage of the various variables encountered during drone flights, such as weather conditions, changes in illumination, variations in background elements, and other environmental factors. To bridge this diversity gap, we developed the initial Long-Range Drone Detection Dataset (LRDD) \cite{rouhi2024long}. This dataset comprises drone images captured in diverse weather conditions, lighting scenarios, and against varying backgrounds, incorporated with a wide range of drone orientations and flight patterns. Benchmark tests utilizing LRDD in drone detection tasks demonstrated a notable improvement in algorithm performance compared to those trained without LRDD.
Despite its contributions, further enhancements were required to address additional gaps in the dataset and enhance the performance of drone detection algorithms. Thus, we introduce LRDDv2, a significant augmentation of the original dataset. LRDDv2 stands out as it includes drone range data for over 8,000 images, a feature not present in any existing dataset. The inclusion of drone range information in LRDDv2 enables supervised training of vision algorithms capable of estimating target range accurately. The majority of images in LRDDv2 depict drones captured at long ranges.  Additionally, LRDDv2 provides comprehensive weather conditions, timestamps indicating the time of day, and a diverse array of background environments. The dataset's incorporation of small drone images enhances performance in long-range detection tasks, while the dataset's variations ensure more robust target detection overall. Some specific aspects of dataset diversities are addressed as follows:

\textbf{Occlusion and Blending into the Background:} UAVs often operate in environments where they can easily blend into the background or be partially or fully occluded by objects, such as buildings, trees, or other structures. Addressing this challenge requires datasets that include scenarios where drones are not immediately distinguishable from their surroundings, testing the algorithms' ability to detect UAVs under compromised visibility conditions.

\textbf{Illumination Variation:} The appearance of drones can significantly change with varying lighting conditions. Datasets must include images captured at different times of the day and under various lighting conditions, such as dawn, dusk, and noon, to train algorithms to recognize drones regardless of the illumination level and shadow effects.

\textbf{Weather Conditions:} Weather plays a critical role in UAV visibility and detectability. LRDDv2 incorporates a wide range of weather conditions, from clear skies to overcast and rainy scenarios, challenging algorithms to maintain detection performance despite adverse weather effects. 

\textbf{Diverse Background:} UAVs are deployed in various settings, from urban to rural and natural landscapes. Each environment presents unique challenges, such as different background textures, colors, and complexities. By including images from diverse backgrounds, LRDDv2 aims to improve algorithms' adaptability and accuracy across multiple operational contexts.

While LRDDv2 addresses numerous challenges essential for a comprehensive drone detection dataset, its two primary novelties lie in 1- capturing drones at long distances and 2- providing explicit range information for the drones within the scene. Through the detailed examination of these challenges and the presentation of LRDDv2's comprehensive approach to addressing them, this paper aims to contribute significantly to the field of UAV surveillance and safety, enabling more reliable and efficient integration of drones into national and global airspaces.

\section{Related Works-Existing Datasets}

Various datasets have been curated to facilitate the training of computer vision models aimed at object detection and recognition within the scope of Unmanned Aerial Vehicles (UAVs). These existing datasets predominantly utilize imagery captured by drone-mounted cameras and aim to improve the autonomous capabilities of UAVs in various environments. In this section, we will assess the available datasets and evaluate their strengths and weaknesses.

The \textbf{Drone vs. Bird challenge} dataset \cite{coluccia2021drone} includes annotated videos featuring drones and birds, created inaugural round of the challenge. The dataset contains 77 video sequences captured using MPEG4-encoded cameras. These videos were obtained utilizing both fixed and mobile camera setups and feature eight different drone models. While comprising imagery that addresses occlusion, it does not account for variability in lighting and meteorological conditions.

The \textbf{UG2} dataset \cite{vidal2018ug} consists of video footage from UAVs, manned gliders, and ground cameras captured under challenging conditions. Although it contained 37 different classes, it includes "Aircraft" as a class but does not have "drone" as one. The dataset is suitable for detection and tracking and addresses challenges like lens flares and unstable camera positions, although it does not cover variations in lighting.

The \textbf{DAC-SDC} dataset \cite{xu2019dac} provides a collection of 150,000 images acquired via DJI drones. These images are manually annotated into 12 primary categories alongside 95 sub-categories. Notably, 'drone' is one of the main categories, further divided into four distinctive sub-categories. Despite this focus on drones, the dataset neglects variables such as alterations in lighting, obstacles, or weather variations.

The \textbf{Real World Object Detection Dataset for Quadcopter UAV Detection} (RWOQ) \cite{pawelczyk2020real} encompasses drones in a variety of types, sizes, positions, and environments. The dataset is consists of 56,821 images auto extracted from RGB videos and is divided into two distinct "drone" and "no-drone" categories. It confronts several difficulties, including differing weather situations and lighting conditions, along with occasional occlusions. However, a substantial portion of the images predominantly display drones against clear and unobstructed skies.

The \textbf{U2U-D\&TD} dataset \cite{li2021fast} contains 50 video sequences showcasing the flight of multiple UAVs, recorded using a GoPro camera mounted on a single UAV. The dataset provides high-definition resolutions and features real-world challenges such as illumination variation, background clutter, and small target objects. Although it encompasses occluded targets and images with different lighting conditions, it does not account for changes in weather conditions.

The \textbf{Airborne Object Tracking} (AOT) dataset \cite{airborne_object_tracking_dataset} comprises 4,943 flight sequences, averaging about 120 seconds per sequence. With over 5.9 million images and 3.3 million accompanying annotations, the dataset was created by equipping two aircraft with sensors and flying planned encounters. While this dataset includes various sky conditions, such as partial cloudiness, it fails to offer variety regarding additional weather variations like rain and snow, and diverse illumination conditions.

The \textbf{VisioDECT} dataset \cite{n27q-7e06-22} is designed for the detection and classification of various types of drones. It encompasses 20,924 sample images and corresponding annotations, collected from six different drone models. The dataset includes images of three distinct weather conditions: sunny, cloudy, and evening. It also features variations in altitudes and distances ranging from 30m to 100m. However, it lacks ground truth information regarding object distance, similar to other datasets. Although the dataset addresses illumination and weather variations, it does not include scenarios with occlusion.

The \textbf{DetFly} dataset (also referred to as YOLOv8-DetFly-02) \cite{yolov8-detfly-02_dataset} was released in 2023, emphasizing drone detection, providing a collection of 6,891 images. Although it features a variety of camera angles, lighting conditions, and occlusion scenarios, it does not specifically consider different weather conditions. Again in 2023, the \textbf{UAV Detect} dataset \cite{uav-detect-pfiqs_dataset} was published aiming to detect drones in various background settings. The dataset consists of 20,400 images, showcasing different backgrounds such as sky, forest, and city. While the dataset accounts for illumination variation, it does not provide scenarios with occlusion and weather variations.

The \textbf{Drone Hunter} dataset \cite{wyder2019autonomous} is designed for the detection of small UAVs in GPS-denied environments. The dataset consists of 58,647 annotated images collected and generated within indoor environments, using GoPro cameras. Although it provides a variety of diverse indoor backgrounds and illumination conditions, the dataset does not include outdoor images, variations in weather conditions, and scenarios with distant targets. 

The \textbf{MCL Drone} dataset \cite{chen2017deep} is composed of two parts. The first part contains 30 YouTube video sequences, captured in an indoor or outdoor environment, with some videos featuring multiple drones. The second part of the dataset consists of 30 video clips, shot with a single drone model. Although the dataset includes variations in backgrounds, camera angles, and some illustration conditions, it does not address variety regarding weather variations and scenarios with occlusion.

The \textbf{Multi-Sensor Drone Detection} dataset \cite{svanstrom2021real} features visible and infrared videos, and provides annotations for 4 areal objects: Airplane, Bird, Drone, and Helicopter. The dataset encompasses 650 videos (285 visible and 365 infrared) with a duration of 10 seconds. It also includes a sensor-to-target distance that is divided into three distance category bins: Close, Medium, and Distant. While this dataset provides some weather variations, it fails to offer variety regarding additional weather variations like rain and snow, occlusion scenarios, and lighting conditions.

 The \textbf{Multi-View Drone Tracking} dataset \cite{li2020reconstruction} consists of 5 scenarios designed for capturing 3D drone trajectories from multiple unsynchronized cameras. It employs a variety of different cameras recording the same drone from multiple stationary angles. The dataset only provides the drone’s location in each frame instead of bounding boxes. It includes sunny and cloudy weather conditions, but it predominantly features sky as background, lacking variations in the environment and scenarios involving occlusion.

The \textbf{Anti-UAV} dataset \cite{jiang2021anti} includes visible and infrared dual-mode information, featuring 318 fully annotated videos. The dataset aims to address the issue of vision-based detectors performing poorly in night, as the recorded videos contain two categories of day and night environments. Despite its strengths, the dataset lacks location distribution and the drones primarily concentrated in the central area. Additionally, it fails to provide various illumination conditions, scenarios with occlusion, and different weathers.

The \textbf{DUT Anti-UAV} dataset \cite{zhao2022vision} comprises two parts focusing on drone detection and tracking, showcasing various types of UAVs. It includes 10,109 images for the detection part, and 20 video sequences (totaling 24,804 images) for the tracking section. While the dataset offers some variations in weather conditions and backgrounds, it predominantly features images with drones at close range and lacks images with distant targets.

Previously, we published the \textbf{Long-Range Drone Detection} (LRDD) dat\-aset \cite{rouhi2024long}, a comprehensive training dataset aimed to alleviate the scarcity of practical drone detection datasets. This dataset encompassed a set of different UAV types, flight patterns, variations in altitude, and environmental conditions. It is also suitable for urban development due to its inclusion of different backgrounds and distant targets. The dataset consists of 22,500 images having a resolution of 1920x1080 pixels, and depicts multiple drones within certain frames. As the dataset was intended to capture videos of drones at substantial distances, it led to image frames where the drones occupied 50 or fewer pixels. It also showcases various weather conditions, illumination variations, and scenarios with occlusion and drone blending with the background.

Existing datasets for object detection using drone images pose limitations, including limited volume, insufficient variety in terms of lighting, climatic conditions, backgrounds, camera angles, and blending. Additionally, many of these datasets lack long-range images featuring smaller drones. The primary concern lies in the absence of object range information crucial for drone collision avoidance, severely limiting the training of computer vision algorithms for autonomous operations.
\begin{table}[htbp]
\centering
\caption{Overview of Drone Detection Datasets}
\label{tab:drone_datasets}
\tiny
\begin{tabular}{|l|c|c|c|c|c|}
\hline
\textbf{Dataset} & \textbf{Objects} & \textbf{Illumination} & \textbf{Occlusion} & \textbf{Weather} & \textbf{Providing Range} \\ 
\hline
Drone vs. Bird & Drone + Bird & No & Yes & No & No \\ \hline
UG2 & Aircraft (no Drone) & Yes & No & No & No \\ \hline
DAC-SDC & Drone & No & No & No & No \\ \hline
RWOQ & Drone & Yes & Partial & Yes & No \\ \hline
U2U-D\&TD & Drone & Yes & Yes & No & No \\ \hline
AOT & Aircraft (no Drone) & No & No & Partial & No \\ \hline
VisioDECT & Drone & Yes & No & Yes & No \\ \hline
DetFly & Drone & Yes & Yes & No & No \\ \hline
UAV Detect & Drone & Yes & No & No & No \\ \hline
Drone Hunter & Drone & Partial & No & No & No \\ \hline
MCL Drone & Drone & Partial & No & No & No \\ \hline
Multi-Sensor & Multiple (incl. Drone) & No & No & Partial & Categorical (Close, Medim, Distance) \\ \hline
Multi-View & Drone & No & No & Partial & No \\ \hline
Anti-UAV & Drone & Partial & No & No & No \\ \hline
DUT Anti-UAV & Drone & No & No & Partial & No \\ \hline
LRDDv1 & Drone & Yes & Yes & Yes & No \\ \hline
\end{tabular}
\end{table}
\section{Long Range Drone Detection (LRDD)v2 Dataset}
The LRDDv2 dataset, comprising 39,516 images, represents a significant advancement in drone detection, offering diverse environments and scenarios.  Among these, \textbf{8,000 annotated images include range information}, enhancing the dataset's utility for developing models capable of estimating distances to drones. Distinctive in its emphasis on long-range drone imagery, these images are crucial for the development of detection models that effectively identify drones at great distances, where they appear as mere specks against vast backgrounds. In continuation, we enumerate the most significant challenges that the presented dataset covers:

\textbf{Long Distance Images}
The dataset features a substantial number of images capturing drones up to distances of 350 ft from the camera. This inclusion of long-range imagery is vital for enhancing prediction models specifically designed for distant UAV detection, a key requirement for real-world applications. Such diverse distance coverage ensures that detection systems trained on this dataset are well-prepared for practical surveillance challenges. An illustrative example of long-distance drone imaging within this dataset is presented in Figure \ref{fig:long_disance}.

\textbf{Moving Camera}
Reflecting realistic operational conditions, the dataset incorporates sequences where not only the drone is in motion, but the camera is also moving. This addition presents a real-world challenge that enables the training of more robust detection models. 

\textbf{Occlusion}
LRDDv2 contains a variety of cases where drones are partially or fully occluded by objects such as trees or bridges. Figure~\ref{fig:occluded} displays an instance of a drone that is partially obscured by tree leaves.

\textbf{Background Blending}
Our dataset includes a range of images where drones blend into the background. Figure~\ref{fig:blend_park} demonstrates this challenge with a drone operation in an urban park setting, where the UAV's appearance is congruent with the environment, necessitating advanced pattern recognition for detection.

\textbf{Illumination Challenges}
Collected under different lighting conditions, from bright daylight to the low light of the evening, the dataset also tackles common issues like glare. Figure~\ref{fig:glare} exemplifies such a scenario where glare affects the visibility of the drone.

\textbf{Weather Conditions}
The dataset features images taken in various weather conditions, enhancing the robustness of detection under different meteorological scenarios. The weather categories are as follows: 1) Sunny, 2) Clear, 3) Cloudy, and 4) Rainy.  Figure~\ref{fig:rural_cloud} illustrates an example captured on a cloudy day.

\textbf{Multiple Object Detection}
In many instances, the dataset presents the challenge of detecting multiple drones within a single scene. Figure~\ref{fig:blend_building} highlights this by showing two drones, with one drone blending into the urban architecture.

\begin{figure}
  \centering
\includegraphics[width=0.65\textwidth]{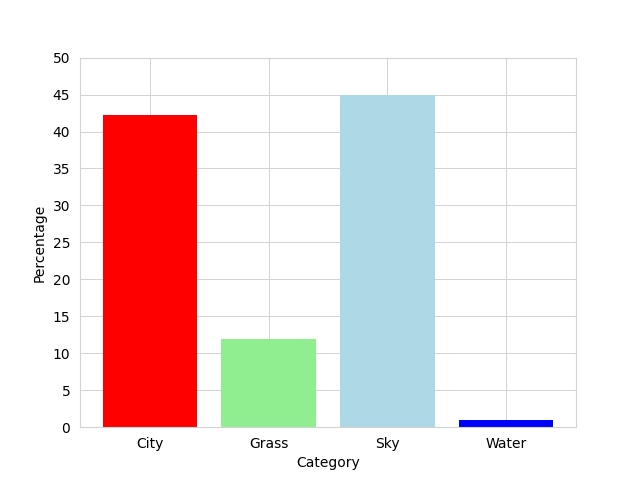}
  \caption{Distribution of Background by percentage}
  \label{fig:sta_back}
\end{figure}

\begin{figure}
  \centering
\includegraphics[width=0.9\textwidth ,height=0.5\textwidth,keepaspectratio]{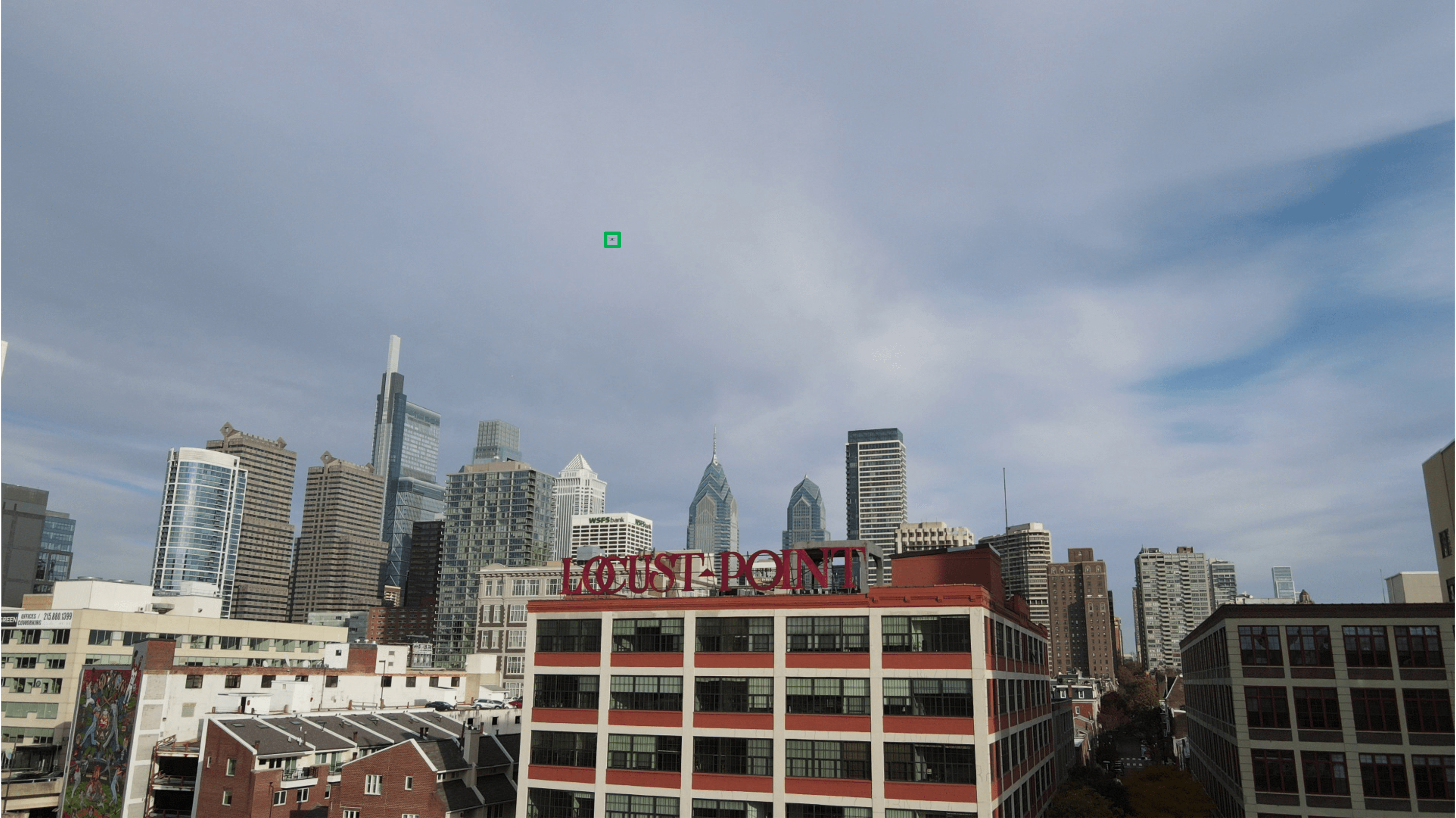}
  \caption{Example of a drone captured at a long distance}
  \label{fig:long_disance}
\end{figure}

\begin{figure}
  \centering
\includegraphics[width=0.9\textwidth ,height=0.5\textwidth,keepaspectratio]{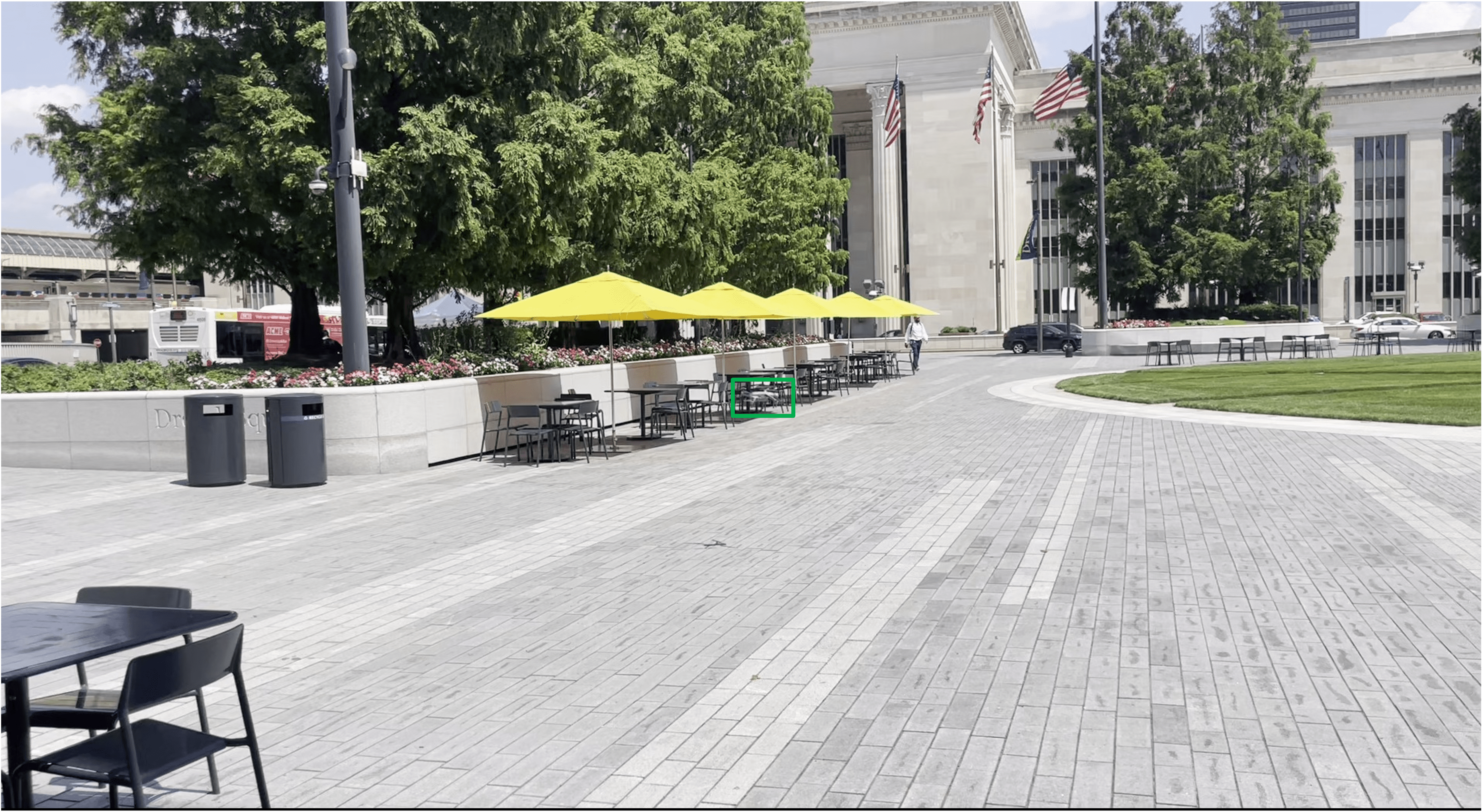}
  \caption{Example of UAV blending into an urban park environment, illustrating the challenge of detecting drones against complex backgrounds within the LRDDv2 dataset.}
  \label{fig:blend_park}
\end{figure}

\begin{figure}
  \centering
\includegraphics[width=0.7\textwidth ,height=0.5\textwidth,keepaspectratio]{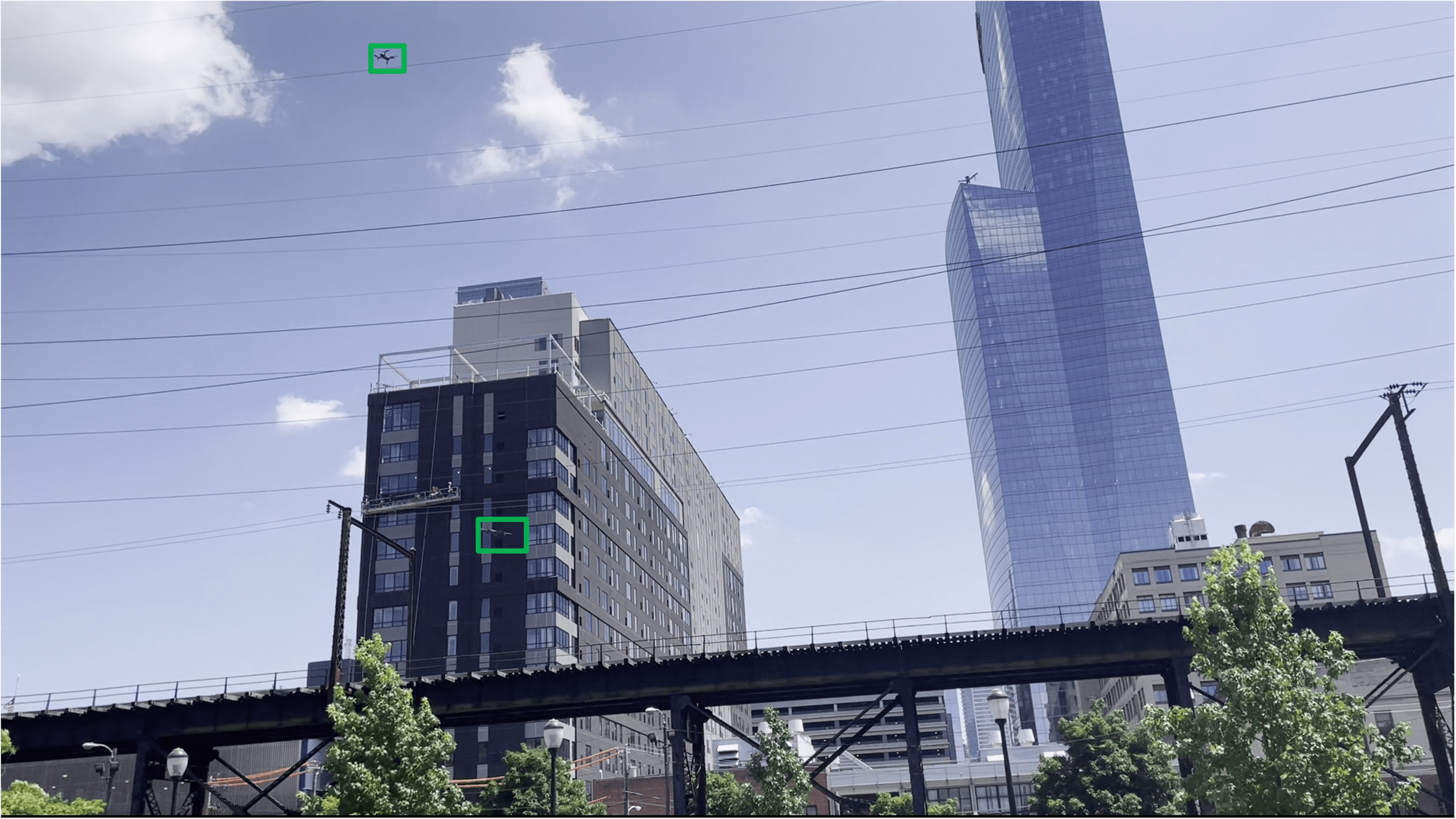}
  \caption{Example of Complex UAV Detection Scenario from LRDDv2: Aerial view showing two drones, with one exhibiting near-perfect background blending against an urban structure, posing a significant challenge for detection algorithms}
  \label{fig:blend_building}
\end{figure}

\begin{figure}
  \centering
\includegraphics[width=0.7\textwidth ,height=0.5\textwidth,keepaspectratio]{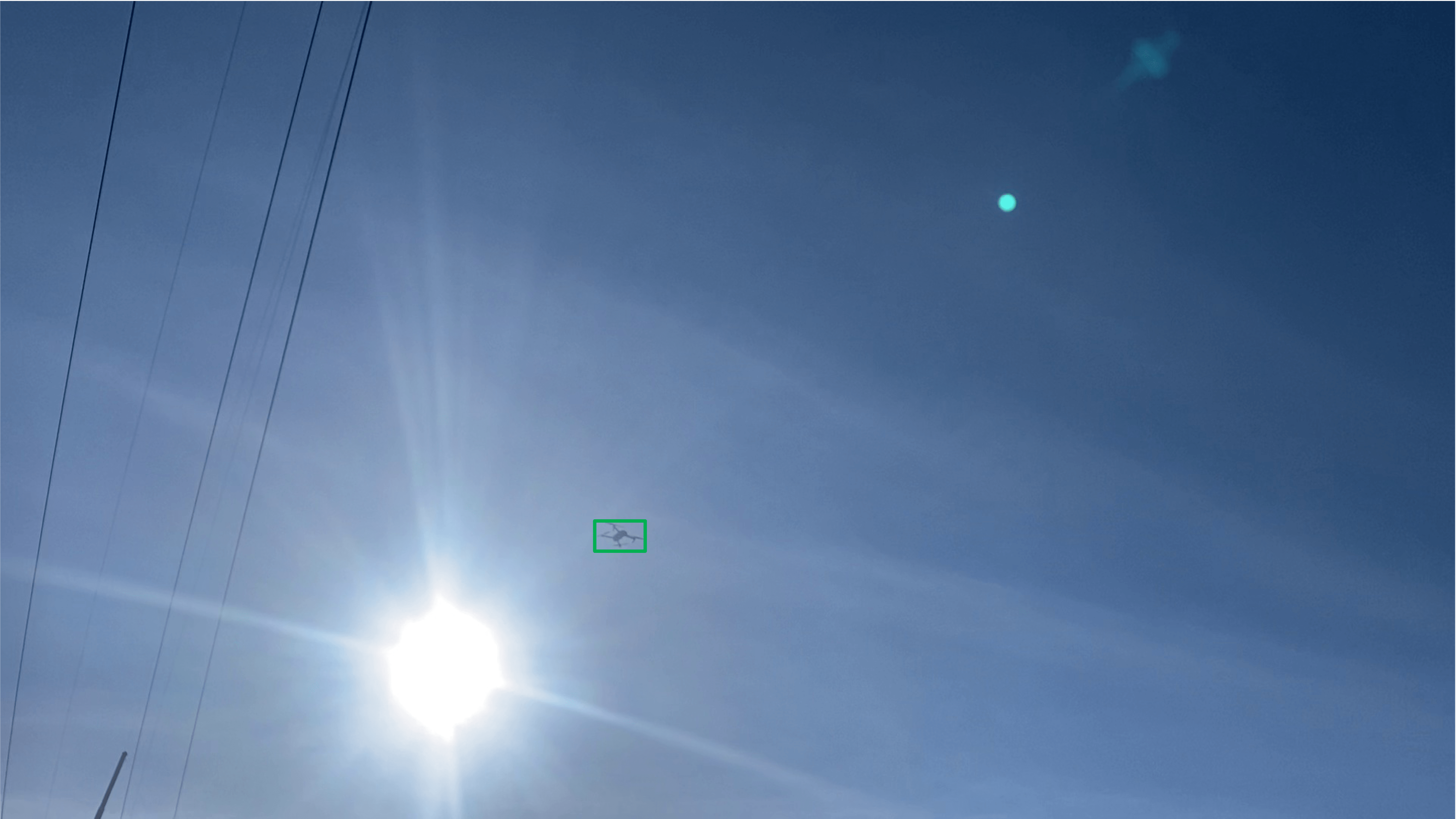}
  \caption{Example of a drone image featuring a glare challenge }
  \label{fig:glare}
\end{figure}

\begin{figure}
  \centering
\includegraphics[width=0.7\textwidth,keepaspectratio]{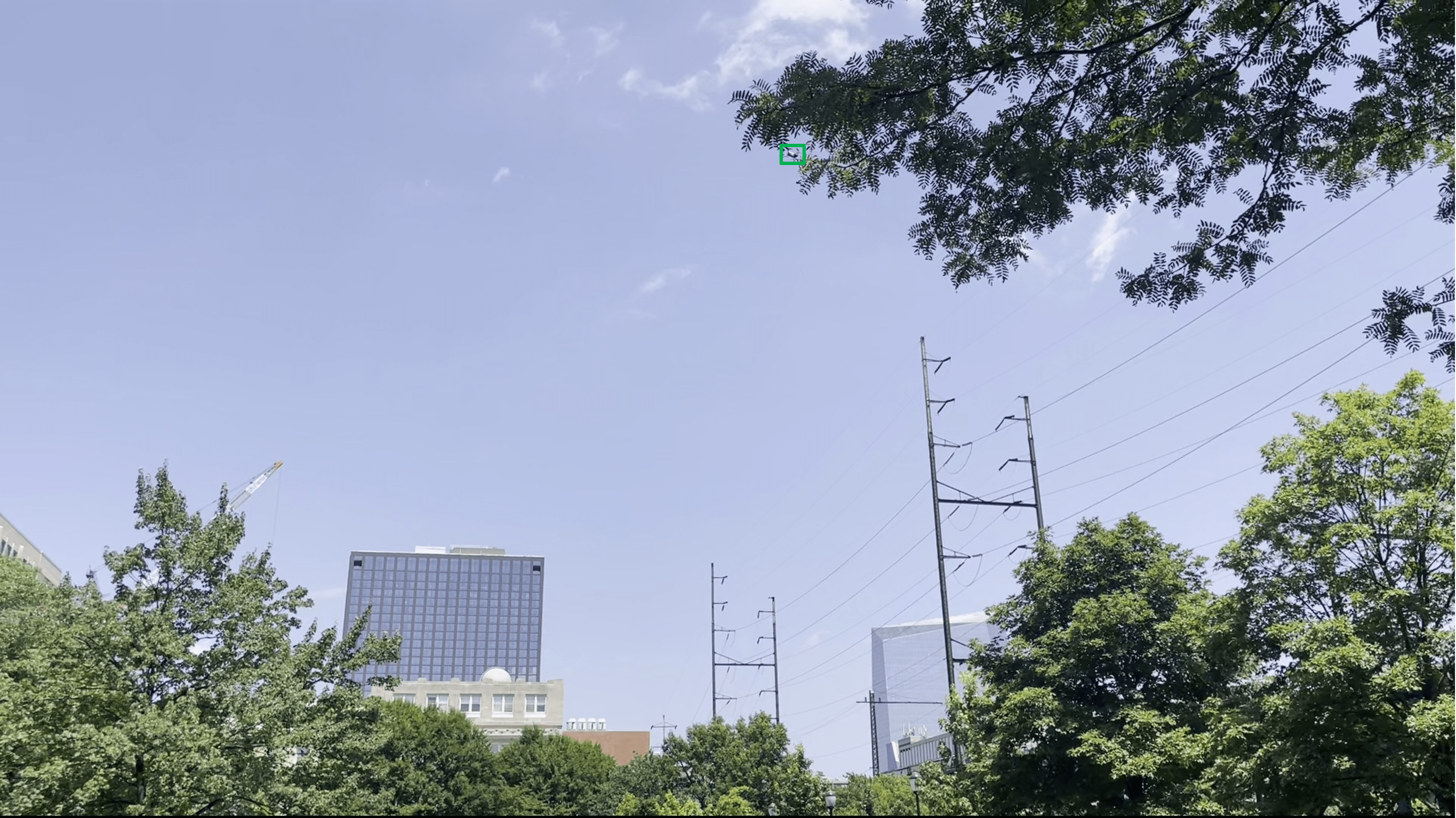}
  \caption{Example of a drone partially obscured by tree leaves}
  \label{fig:occluded}
\end{figure}

\begin{figure}
  \centering
\includegraphics[width=0.85\textwidth ,height=0.5\textwidth,keepaspectratio]{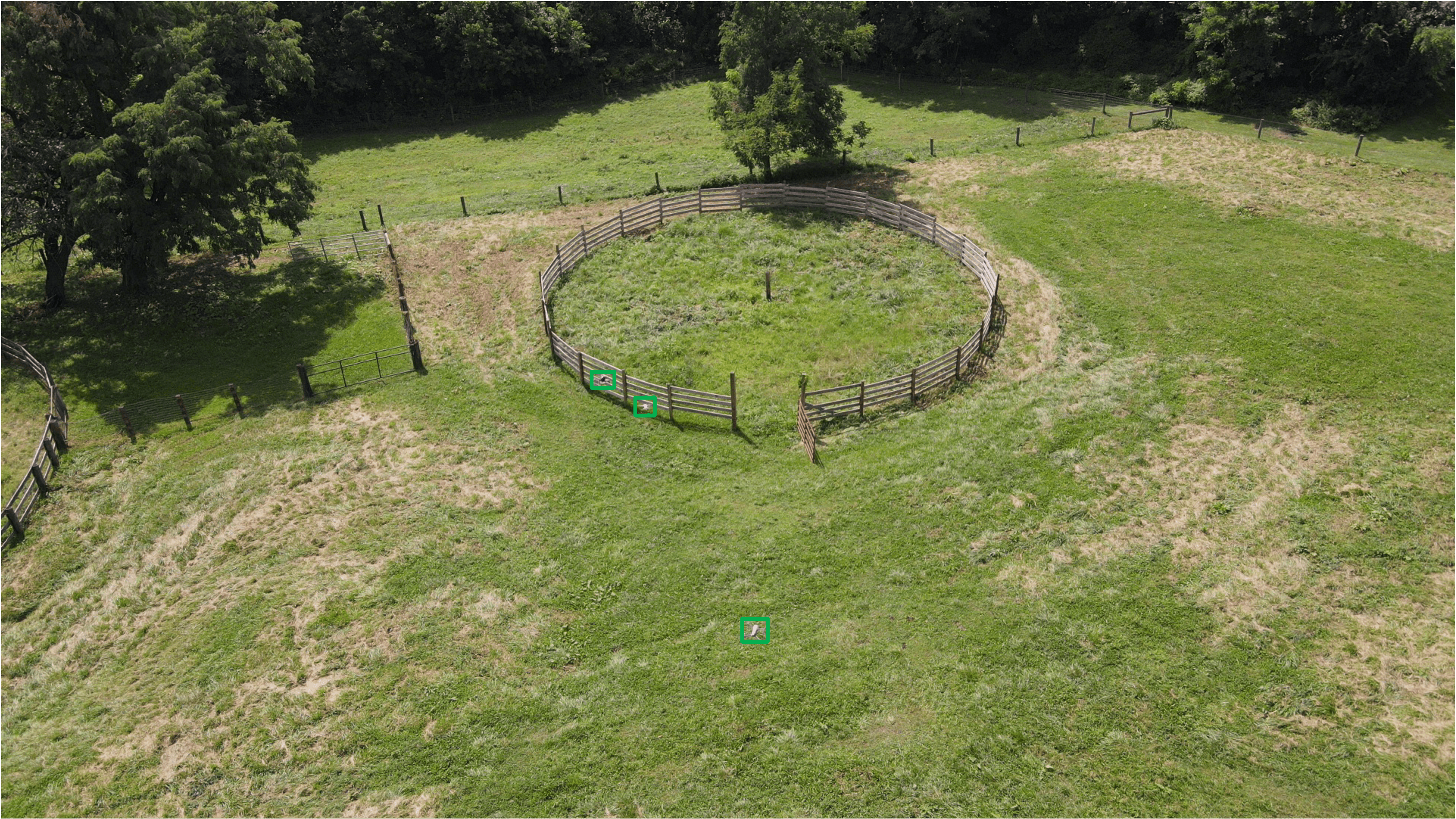}
  \caption{Example showcasing three flying drones in a rural environment, addressing two challenges: diverse surroundings and the presence of multiple drones}
  \label{fig:rural}
\end{figure}

\begin{figure}
  \centering
\includegraphics[width=0.85\textwidth]{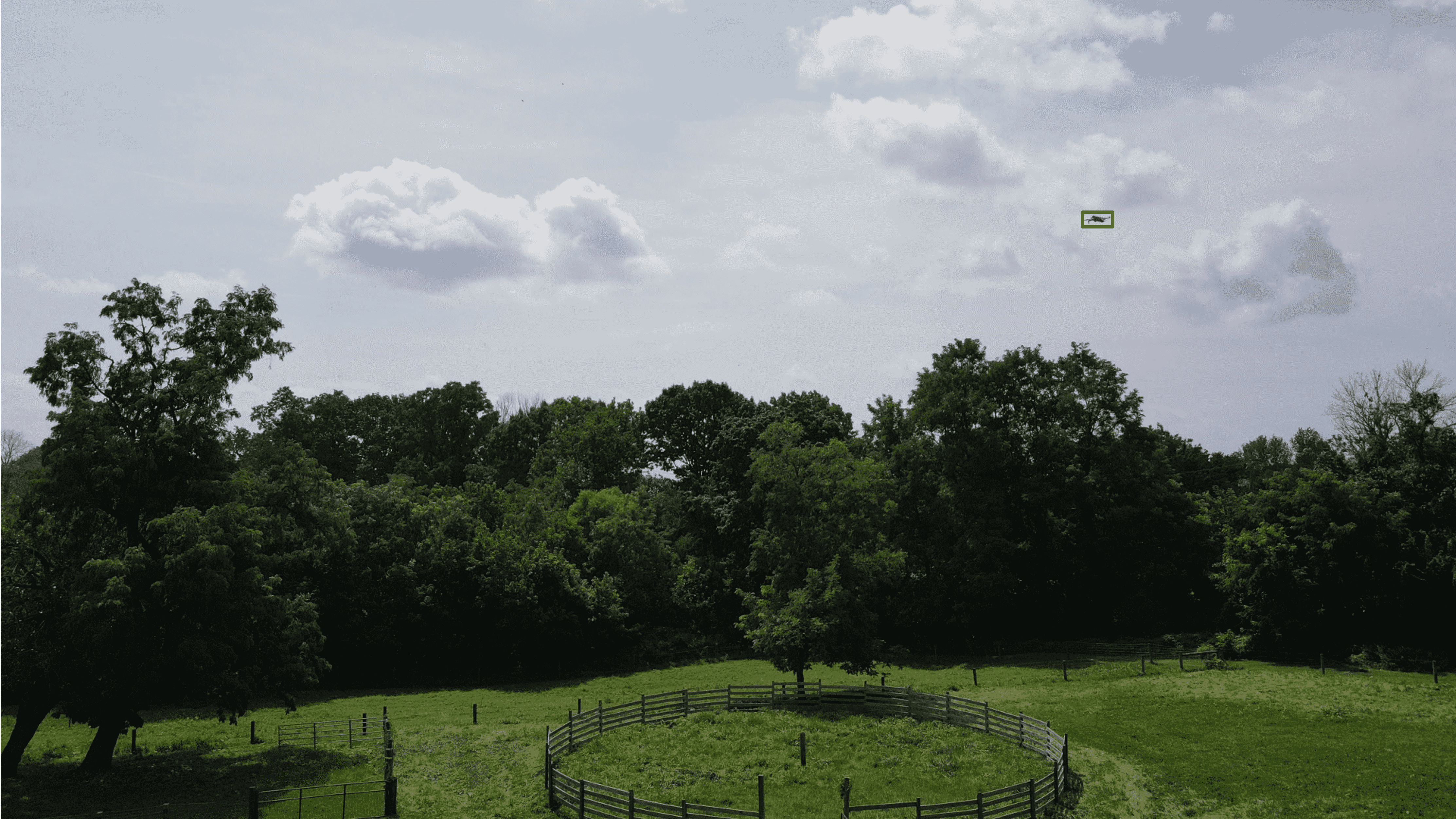}
  \caption{Example of a drone captured in a rural environment under cloudy weather conditions }
  \label{fig:rural_cloud}
\end{figure}

\textbf{Diverse Background}
LRDDv2 extends across different environments, from urban to rural settings, which aids in training models on a wide array of backgrounds. The main background categories are City, Grass, Sky, and Water. Figure 3 displays the background distribution within the proposed dataset \ref{fig:sta_back}. 
Figure~\ref{fig:rural} portrays an example captured in a rural setting that also includes multiple drones, thus addressing two mentioned challenges. 

The LRDDv2 dataset, with its emphasis on long-range images and real-world challenges such as occlusion, background blending, diverse illumination, varying weather conditions, multiple object detection, and a range of background, serves as a comprehensive tool for advancing drone detection technology. It is poised to significantly enhance the performance and reliability of UAV detection algorithms in real-world surveillance and security operations.

%LRDD2:

\subsection{Data Collection}
The images in the dataset were collected using the DJI Mavic Air 2 drone capable of collecting 1080p 30 fps video footage, as well as using cellular cameras (iPhone 12/15 Pro Max and Google Pixel 6) capable of capturing 30 fps 1080p video footage. In both cases the camera collected video footage of other drones, which were the and the DJI Mini 3 and the DJI Mavic Pro. The footage includes many scenarios, including multiple and single drone frames, diverse backgrounds, and a wide range of distances from the camera. 

\subsection{Data Preprocessing}
To extract the images in the dataset, frames from the video footage are sampled at regular intervals of 10 fps. These images are then synchronized with the range information collected from the flight logs of the drones. The range information collected by the drone includes height and distance from the home point where the flight began. In the case of a stationary cellular camera collecting footage at ground level, the absolute distance from the drone is calculated with the following formula:
\begin{equation}
    \text{Distance}_{\text{Absolute}} = \sqrt{\text{Distance}_{\text{Horizontal}}^2 + \text{Height}_{\text{Drone}}^2}
\end{equation}

For aerial footage, both the drone and camera start from the same home position. The horizontal distance is calculated using GPS coordinates (latitude and longitude) of the drone and camera, applying the haversine formula. The absolute distance is then determined using the following equations:
\begin{equation}
    \Delta \text{Height} = \left| \text{Height}_{\text{Target}} - \text{Height}_{\text{Camera}} \right|
\end{equation}

\begin{equation}
    \text{Distance}_{\text{Absolute}} = \sqrt{\text{Distance}_{\text{Horizontal}}^2 + (\Delta \text{Height})^2}
\end{equation}

\begin{figure}
  \centering
\includegraphics[width=0.55\textwidth]{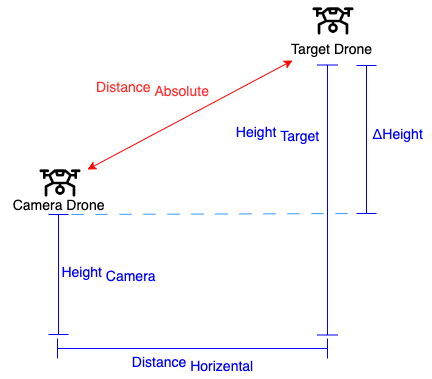}
  \caption{Illustration of distance calculation in the dataset in the case of the footage being collected by an aerial drone.}
  \label{fig:dist}
\end{figure}

Figure \ref{fig:dist} shows an illustration of the distance calculation process, highlighting the relationship between the horizontal distance, height delta, and absolute distance in the case of footage being collected by an aerial drone.

Data was labelled using the YOLO format. Both manual and assisted bounding box generation methods were used. For the \textbf{automated bounding box generation}, a sample of manually labelled images are used as training data for a YOLOv5 algorithm to automatically detect possible matches. This process speeds up labelling and allows for the boxes to merely be tweaked by the labeler. For the \textbf{manual labelling and bounding box correction}, the open-source Python-based software LabelImg [15] was used. This tool provides a simple GUI to create and edit bounding boxes and then automatically generate the labels in YOLO format.

\section{Dataset Benchmarking}

We evaluated the YOLOv8m model for performance improvements brought about by the Long Range Drone Detection (LRDD)v2 data through a strict benchmarking mechanism across a number of datasets that included Drone-vs-Bird \cite{coluccia2021drone} , Detfly \cite{yolov8-detfly-02_dataset} , UAV-Detect \cite{uav-detect-pfiqs_dataset}.  This detailed study focused on the adaption and realization of YOLOv8 model under different datasets and detection scenarios. The benchmarking process focused on two primary metrics: mean Average Precision (mAP) at IoU threshold 0.5 (mAP@50) and mAP measured over IoU thresholds from 0.5 to 0.95 (mAP@50-95), offering details both on the precision of detection at a common threshold as well as a more fine-grained assessment across a range of detection accuracies.

\section{Results and Discussion}

The experimental results presented in Table~\ref{tab:yolo_performance2} show the performance of the YOLOv8m model in detecting drones across different datasets, which include Detfly and UAV-Detect, when trained with different datasets combinations: Drone-vs-Bird, LRDDv2, and both of them. 

As shown in Table~\ref{tab:yolo_performance2}. The model’s performance increases in all measures while it is being trained on the union dataset of Drone-vs-Bird and LRDDv2, yielding the highest mAP@50 and mAP@50-95 scores for both the Detfly and UAV-Detect evaluation datasets. Training the model exclusively on LRDDv2 yields better inference outcomes on both evaluation datasets compared to using the Drone vs Birds dataset alone; however, the most effective results are obtained when training incorporates both datasets together. This enhancement highlights the complementary nature of the Drone-vs-Bird and LRDDv2 datasets. While the Drone-vs-Bird dataset provides a broad range of general drone and bird images, the LRDDv2 dataset introduces specific challenges inherent to drone detection, including long-range detection capabilities, varied lighting and  scenarios involving blending and occlusion.

The improvement in model performance on the Defly dataset in the mAP@50 metric is from 0.376 to 0.463, and for mAP@50-95, it increased from 0.14 to 0.26 when trained on aggregated datasets, allows a conclusion that the model acquired better skills to generalize over diverse scenarios. Equally on the UAV-Detect dataset, training with the two datasets provides a noticeable boost that increases the mAP@50 from 0.510 to 0.644 and the mAP@50-95 from 0.22 to 0.32. The improvements represent the strength brought by the LRDDv2 dataset to the model when dealing with challenging drone detection scenarios.

The results provide a strong evidence of the benefit of using varied training data, especially LRDDv2 dataset that is specifically tailored towards the subtle difficulties in drone detection at long range. The improved performance on both Detfly and UAV-Detect evaluation datasets emphasizes the significance of challenging and comprehensive datasets in the development of drone detection technologies.

\begin{table}[htbp]
\caption{YOLOv8 detection accuracy on Detfly and UAV-Detect using different training datasets.}
\label{tab:yolo_performance2}
\begin{center}
\begin{tabular}{llcc}
\hline
\multicolumn{1}{l}{Training Dataset} & \multicolumn{1}{l}{Evaluation Dataset} & \makecell{mAP@50} & \makecell{mAP@50-95} \\[2pt]
\hline\rule{0pt}{12pt}Drone-vs-Bird     & Detfly       & 0.376           & 0.14         \\
LRDDv2            & Detfly       & 0.458           & 0.27         \\
Drone-vs-Bird+ LRDDv2 & Detfly   & 0.463           & 0.26         \\
Drone-vs-Bird     & UAV-Detect   & 0.510           & 0.22         \\
LRDDv2            & UAV-Detect   & 0.562           & 0.30         \\
Drone-vs-Bird+ LRDDv2 & UAV-Detect & 0.644           & 0.32        \\[2pt]
\hline
\end{tabular}
\end{center}
\end{table}

\section{Detection Probability Versus Bounding box area}

\begin{figure}
  \centering
\includegraphics[width=\textwidth ,height=0.5\textwidth,keepaspectratio]{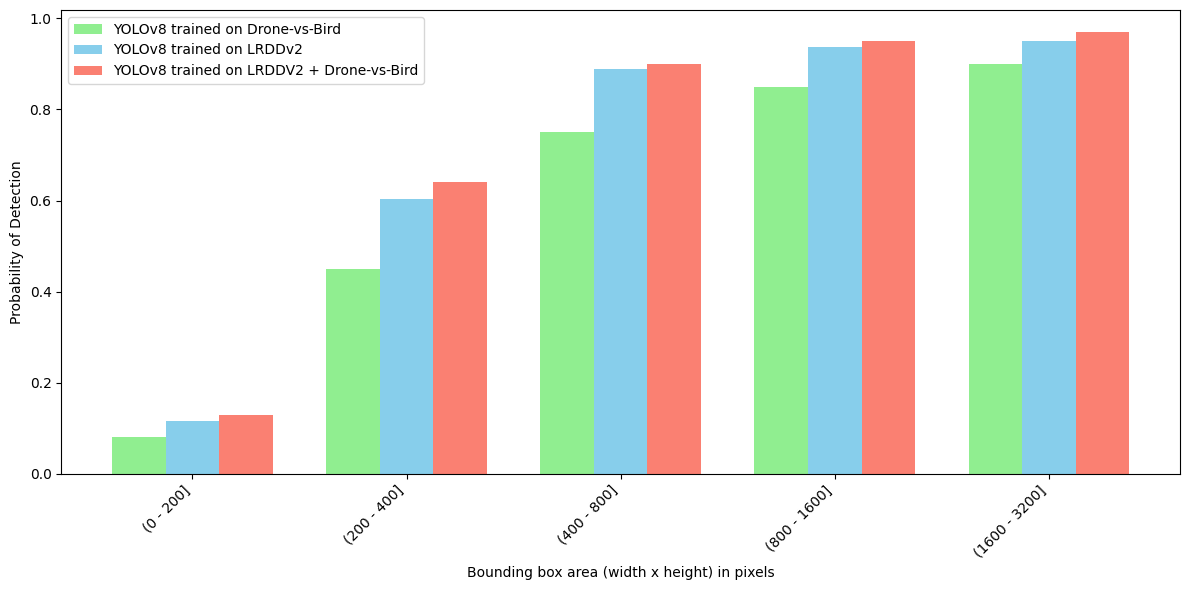}
  \caption{Detection Probability by Bounding Box area (pixel): The graph shows the YOLOv8m model's detection probability as a function of bounding box size.}
  \label{fig:dis-plot}
\end{figure}

An essential component of drone detection algorithm evaluation is the investigation of how detection likelihood correlates with bounding box dimensions, which serve as a proxy for the UAV's range from the capture device. In our study, we leveraged the YOLOv8m model trained on three scenarios: the Drone-vs-Bird dataset, the LRDDv2 dataset, and a combination of Drone-vs-Bird + LRDDv2 datasets, to explore this relationship. Figure \ref{fig:dis-plot} visualizes the probability of detection by bounding box area across these scenarios.

The figure clearly demonstrates a dependency of detection rates on bounding box sizes. As the distance increases (represented by smaller bounding boxes), the detection probability decreases across all training scenarios. YOLOv8 trained on Drone-vs-Bird alone shows the lowest detection probabilities, especially for smaller bounding boxes. In contrast, training on LRDDv2 yields better results, and the highest detection rates are observed when YOLOv8 is trained on the combination of Drone-vs-Bird + LRDDv2, showcasing the complementary strengths of both datasets.

This pronounced decline in detection probability with decreasing bounding box sizes highlights the critical need for datasets tailored to long-range drone detection. The development and refinement of models to improve their long-range detection capabilities is indispensable for addressing the challenges posed by detecting distant UAVs in real-world surveillance applications.

\section{Challenges and Future Work}

The detection of drones including especially at longer ranges although being highly complex is of significant importance for the operational efficiency of surveillance systems. The main problem is in the small, distant object detection among many backgrounds and environmental conditions. Addressing this challenge can be approached from two angles:

First of all, a need of an advanced dataset, for example, LRDDv2 recommended by this study, is highly essential. This kind of dataset with images of drones at far distances with differently illuminated backgrounds and other related issues creates a good basis for training detection models. The diversity of the dataset enables the model to generalize from the different situations, thus, ensuring accurate detection through numerous real world scenarios.

Building on the rich dataset introduced in this paper, the next step towards facilitating long-distance drone detection is the development of new methods or the refinement of existing algorithms to robustly detect small objects at range. Our suggestion for future research directions includes the application of techniques such as Split-Attention High-resolution Inference (SAHI)\cite{akyon2022sahi,obss2021sahi}, not solely at the inference stage but also during training.

By employing a methodical approach that divides images into patches to increase resolution, models can learn detailed local features that are crucial for detecting small objects. Concurrently, by training on whole images, the model can also learn global features, providing a comprehensive understanding of the scene. This dual focus ensures that models are well-equipped to recognize drones based on both their minute details and their context within the larger environment.

Such an approach has the potential to significantly enhance the performance of object detection models like YOLO. By integrating local and global feature learning, future models can offer more refined and accurate detection capabilities, particularly in the challenging domain of long-range drone surveillance.

Another promising avenue for future work is the development of models that can estimate the range of flying objects. Leveraging the range information provided in our dataset, researchers could explore creating models that not only detect but also determine the distance of drones from the camera. This capability would be instrumental in applications requiring precise location tracking of UAVs, such as air traffic control, collision avoidance systems, and surveillance operations.

\section{Conclusion}

In summary, the development  of the Long Range Drone Detection (LRDD)v2 dataset marks a significant advancement in the field of UAV detection, particularly for long-range applications. By addressing the need for a more diverse and extensive collection of drone images at far distances, LRDDv2 fills a critical gap in existing datasets, offering over 39,516 meticulously annotated images with a distinct emphasis on long-range detection and comprehensive real-world conditions.  The dataset's standout feature — the inclusion of explicit range information for over 8,000 images — empowers the development of sophisticated algorithms that excel not only in drone detection but also in precise distance estimation.  While benchmarking reveals that models trained on LRDDv2 outperform those trained solely on the Drone vs Birds dataset, the highest performance is achieved when both datasets are combined for training. This work contributes to the ongoing effort to safely integrate drones into both national and global airspace.

\bibliographystyle{splncs04}
\bibliography{author}

\end{document}